\documentclass{article}

\PassOptionsToPackage{numbers, compress}{natbib} 
\usepackage[final]{neurips_2025}

\usepackage[utf8]{inputenc} 
\usepackage[T1]{fontenc}    

\usepackage{url}            
\usepackage{booktabs}       
\usepackage{amsfonts}       
\usepackage{nicefrac}       
\usepackage{microtype}      
\usepackage{xcolor}         
\usepackage{algorithm}
\usepackage{amsmath, amssymb}
\usepackage[noend]{algpseudocode}
\usepackage{bm} 
\usepackage{multicol} 
\usepackage{array}
\usepackage{geometry}
\usepackage{setspace}       
\usepackage{multirow}      
\usepackage{colortbl}  
\usepackage{graphicx}  
\usepackage{enumitem}
\usepackage{wrapfig}
\usepackage{subcaption} 
\usepackage{pifont}
\usepackage{arydshln}
\usepackage{hyperref}       
\usepackage{tabularx}
\newcolumntype{Y}{>{\centering\arraybackslash}X}
\title{HBLLM: Wavelet-Enhanced High-Fidelity 1-Bit Quantization for LLMs}

\author{%
	Ningning Chen$^*$ \\
	Sun Yat-sen University \\
	\texttt{chennn27@mail2.sysu.edu.cn}
	\And
	Weicai Ye \thanks{Equal contribution.} \\ 
	Sun Yat-sen University \\
	\texttt{cai\_rcy@163.com}
	\And
	Ying Jiang\thanks{ Correspongding Author.} \\   
	Sun Yat-sen University \\
	\texttt{jiangy32@mail.sysu.edu.cn}
}

\begin{document}

\maketitle

\begin{abstract}
We introduce HBLLM, a wavelet-enhanced high-fidelity $1$-bit post-training quantization method for Large Language Models (LLMs). By leveraging Haar wavelet transforms to enhance expressive capacity through frequency decomposition, HBLLM significantly improves quantization fidelity while maintaining minimal overhead. This approach features two innovative structure-aware grouping strategies: (1) frequency-aware multi-parameter intra-row grouping and (2) $\ell_2$-norm-based saliency-driven column selection. For non-salient weights, a shared mean is employed across quantization groups within each frequency band to optimize storage efficiency. Experiments conducted on the OPT and LLaMA models demonstrate that HBLLM achieves state-of-the-art performance in $1$-bit quantization, attaining a perplexity of $6.71$  on LLaMA$2$-$13$B with an average weight storage of only $1.08$ bits. Code  available at: https://github.com/Yeyke/HBLLM.

\end{abstract}

\begin{wrapfigure}{r}{0.45\textwidth}
	\vspace{-\baselineskip} 
	\centering
	\includegraphics[width=0.45\textwidth]{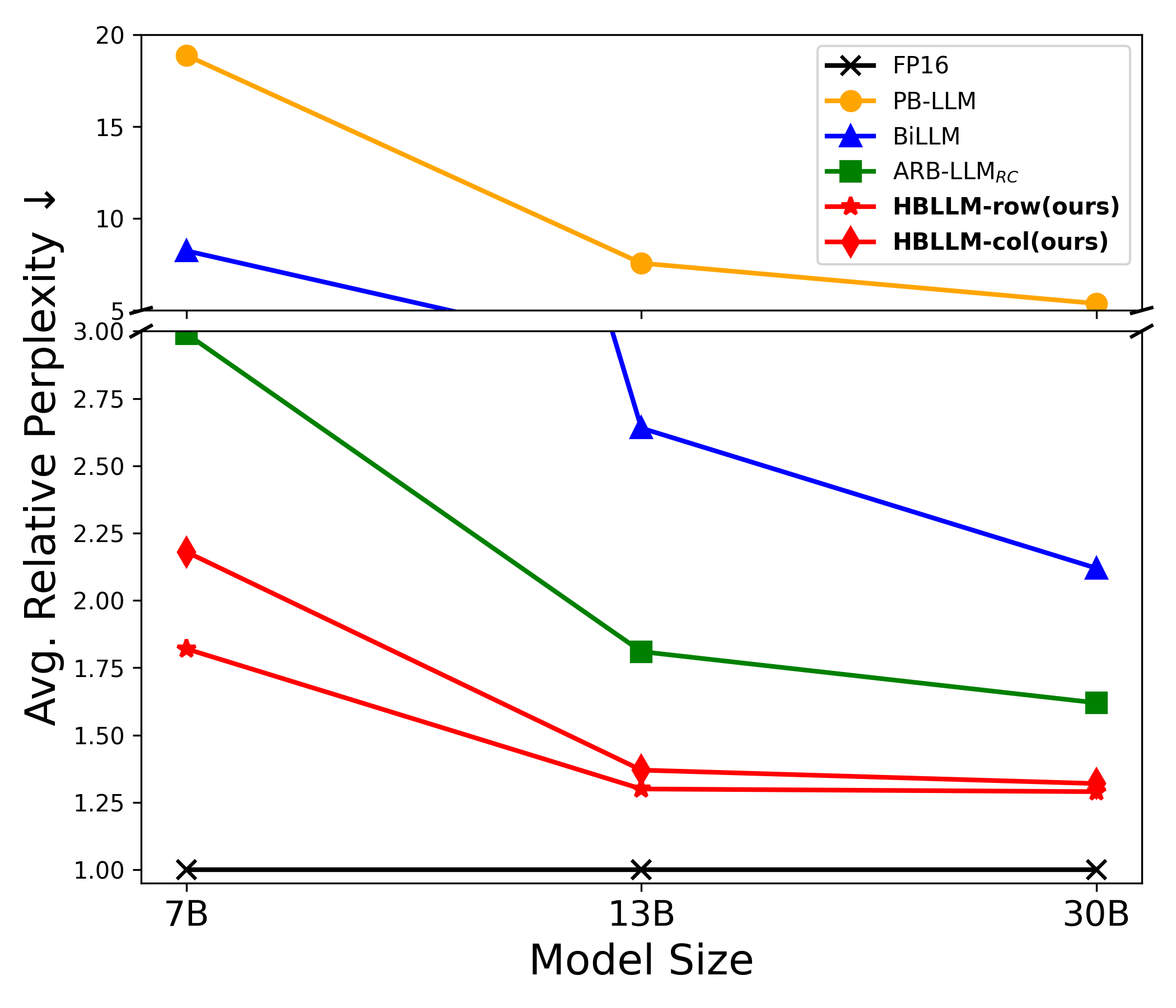}
	\vspace{-14pt} 
	\caption{Average relative perplexity (normalized to FP16) on PTB, WikiText2, and C4 for LLaMA-1 family models, comparing LLM binarization methods and our HBLLM.}
	\label{fig:avgperplexity}
	\vspace{-10pt}  
\end{wrapfigure}

\section{Introduction}

In recent years, Large Language Models (LLMs) have achieved remarkable progress in natural language processing tasks. However, their massive parameter sizes—often reaching tens or even hundreds of billions—pose significant challenges for deployment on edge devices and in low-resource environments. To reduce the computational and memory burden of these models, a variety of compression techniques have been proposed, including 
quantization~\cite{OPTQ2023Frantar,Smoothquant2023Xiao,Zeroquant2022Yao},  pruning~\cite{SparseGPT2023Frantar,ASimple2023Sun}, and knowledge distillation~\cite{Less2023Liang,Distilling2023Shridhar}. Among them, Post-Training Quantization (PTQ) is widely adopted for its efficiency, requiring no additional training and having low deployment cost, especially in $1$-bit quantization, which is considered a key approach for achieving extreme inference efficiency~\cite{survey2024Gong}.

Although existing $1$-bit PTQ methods~\cite{BiLLM2024Huang,Mixture2024Jo, OneBit2024Xu} have achieved some success on base models such as GPT-$2$ and OPT, they tend to suffer from significant performance degradation—or even complete failure—when applied to more complex modern architectures like LLaMA3-8B~\cite{empirical2024Huang}. To address this, recent studies have introduced several strategies to improve quantization fidelity:
\begin{itemize}[leftmargin=1em]
	\item \textbf{Group quantization}: divides the weight matrix into multiple groups for separate quantization. For instance, outlier-aware partitioning handles critical columns independently but can be constrained by partition design and scalability~\cite{BiLLM2024Huang};
	\item \textbf{Residual approximation}: adds residual terms on top of primary quantization to partially recover errors~\cite{Dbllm2024Chen}, though this provides limited fidelity gains and introduces extra computation;
	\item \textbf{Low-Rank Adaptation (e.g., LoRA)}: inserts low-rank modules to absorb quantization errors with some flexibility, like~\cite{ OneBit2024Xu}, but often shows sensitivity to hyperparameters;
	\item \textbf{Global orthogonal transformations}:  apply global rotations in~\cite{FrameQuant2024Adepu,SliceGPT2024Ashkboos,QuIP2023Chee}   before model compression to enhance representational capacity, but require expensive inverse transforms (e.g., matrix multiplications at $\mathcal{O}(d^2)$ complexity for a $d$-dimension linear layer), leading to increased inference latency and energy consumption, making them impractical for deployment.
\end{itemize}

To overcome the structural trade-off between expressiveness and efficiency, we propose a novel $1$-bit PTQ framework—\textbf{HBLLM}. This method is the first to integrate localized orthogonal transformations (i.e., Haar wavelets) into a BiLLM-style quantization process. Combined with structure-aware grouping, HBLLM significantly enhances expressive power under ultra-low bit budgets while maintaining negligible inverse transform cost and excellent compatibility with hardware-efficient inference.

Our main contributions are as follows:
\begin{itemize}[leftmargin=1em]
	\item A \textbf{localized orthogonal transformation mechanism}: we apply a single Haar wavelet transform to decompose the weight matrix into high- and low-frequency components, improving binary expressiveness while reducing transform computation;
	\item \textbf{Frequency-aware multi-parameter intra-row grouping}: we introduce intra-row grouping in the frequency domain to capture structural patterns; 
	\item \textbf{$\ell_2$-norm-based saliency-driven column selection}: we propose an $\ell_2$ norm-based ranking method to retain key columns using saliency metrics, effectively reducing quantization error; 
	\item \textbf{Intra-frequency-band mean sharing}: for non-salient components, we introduce a mechanism that shares the mean across groups within the same row and wavelet band, reducing storage without sacrificing fidelity.
\end{itemize}

We conduct extensive experiments on OPT~\cite{Opt2022Zhang}, LLaMA family~\cite{Llama2023Touvron} of LLMs. Results show that HBLLM achieves state-of-the-art performance under $1$-bit quantization:
Across language modeling tasks (C4, PTB, WikiText2), the perplexity ratio between HBLLM and the original FP16 model remains within the range of $1.2$–$2.2$, shown in Fig~\ref{fig:avgperplexity}, outperforming the next-best methods by $33\%$–$66\%$;
On $9$ zero-shot QA benchmarks, HBLLM retains $73.8\%$–$88.8\%$ of the original model’s accuracy;
On modern architectures such as LLaMA3-8B, HBLLM remains stable with no performance collapse;
Even with a lower average bit rate {and memory usage than BiLLM and ARB-LLM\textsubscript{RC}}~\cite{Arbllm2024Li}, HBLLM outperforms both in overall task accuracy.

These results demonstrate that HBLLM significantly extends the applicability of $1$-bit quantization, balancing extreme compression with high fidelity, and offers a new paradigm for deploying large-scale language models efficiently.

\section{Related Work}

\subsection{1-Bit Post-Training Quantization}
$1$-bit PTQ has emerged as a critical promising solution for deploying LLMs under extremely low bit budgets. Representative methods such as BiLLM~\cite{BiLLM2024Huang} adopt a salient column separation mechanism, in which salient weights are quantized independently, while non-salient weights are grouped based on magnitude and quantized row-wise. ARB-LLM\textsubscript{X}~\cite{Arbllm2024Li} further introduces column-wise grouping and alternating refined binarization, achieving notable improvements in fidelity. Unlike~\cite{OBC2025Edalati,OneBit2024Xu}, BiLLM can accomplish PTQ tasks without intensive computation for knowledge distillation with multi-GPUs.

However, current methods face several key limitations:  
(1) They heavily rely on fixed thresholds or simple $\ell_1$-based heuristics for salient column selection, which are insufficient to capture sparse but significant activation outliers;
(2) They fail to account for the structural asymmetry between row and column dimensions in weight matrices, limiting their adaptability to complex model architectures;
(3) They completely neglect frequency-domain information.

\subsection{Evolution and Limitations of Grouping Strategies}
To improve quantization flexibility and fidelity, some studies have proposed learnable or adaptive grouping strategies. For example, Mixture of Scales \cite{Mixture2024Jo} introduces a Mixture-of-Experts (MoE) mechanism to assign scaling factor groups, and OneBitGPT \cite{OneBit2024Xu} uses frequency masks to control quantization range sensitivity, {and AWQ~\cite{AWQ2023Lin} identifies weights with the greatest impact on model predictions only based on activation outputs}. However, these methods are generally effective only on unstructured tensors, rely on fine-grained distillation, and lack explicit frequency-domain awareness.

In addition, existing grouping strategies \cite{BiLLM2024Huang,Arbllm2024Li} often apply uniform partitioning rules across the entire weight matrix, ignoring variations across different rows. This can lead to degraded expressiveness when quantizing models with significant inter-row diversity.

\subsection{Comparison Between Global Orthogonal Transforms and Local Wavelet Transforms}
Orthogonal transforms have recently been adopted to improve LLM quantization. FrameQuant \cite{FrameQuant2024Adepu} and QuIP \cite{QuIP2023Chee} utilize orthogonal transforms to enhance fidelity, but inference with such global transforms incurs high overhead, requiring $\mathcal{O}(d^2)$ matrix multiplications \cite{FrameQuant2024Adepu}  that cannot be fused into linear layers, leading to increased latency and energy cost.

By contrast, local orthogonal transforms such as the Haar wavelet~\cite{wavelet1999Mallat} offer localized spectral sensitivity and have been widely applied in image compression, denoising, and edge detection~\cite{SAR2017Duan,Waveletsrnet2017Huang}. They can be efficiently implemented via lightweight local convolutions with negligible inference cost, making them well-suited for low-bit compression and edge deployment.

\section{HBLLM: A Quantization Framework with Wavelet Transform and Frequency-Domain Grouping}

\subsection{Motivation and Core Challenges}
Current mainstream $1$-bit quantization methods face three key challenges in practice: 
(1) limited numerical expressiveness leading to high reconstruction error; 
(2) insufficient accuracy in salient column selection, failing to capture critical activation columns;
(3) lack of structure-aware grouping strategies that adapt to heterogeneous model structures.

To characterize expressiveness under ultra-low bit settings, we introduce a new metric: the \textit{cardinality of the Inverse Quantization Set (CIQ)}, which measures the size of the discrete set of dequantized values within a row. CIQ serves as a unified indicator of how the above challenges constrain model fidelity. It acts both as a theoretical tool to analyze the limits of existing methods and as empirical evidence of the advantage of our proposed method.

Under $1$-bit quantization, the CIQ of BiLLM and ARB-LLM\textsubscript{X} is $8$ and $10$, respectively. When block size sets to $128$, the CIQ upper bound of ARB-LLM\textsubscript{X} can reach $128$. In contrast, our method achieves a CIQ of up to $1024$ after applying the Haar wavelet transform, significantly improving theoretical expressiveness. For more information on the benefits introduced by applying Haar transform, please refer to the { appendix B and C. }

Based on aboved analysis, we propose:  
(1) Haar wavelet transform to enhance expressive capacity by frequency decomposition;  
(2) $\ell_2$-norm-based saliency-driven column selection to prioritize critical columns;  
(3) frequency-aware multi-parameter intra-row grouping to capture structural patterns.  
We also introduce an intra-frequency-band mean sharing strategy and local convolution optimization to reduce storage and inference cost, thus forming a $1$-bit PTQ  framework \textbf{HBLLM} .

\begin{figure}[t]
	\setlength{\textfloatsep}{3pt plus 5pt minus 0pt}
	\centering
	\vspace{-8pt}
	\includegraphics[width=0.95\textwidth]{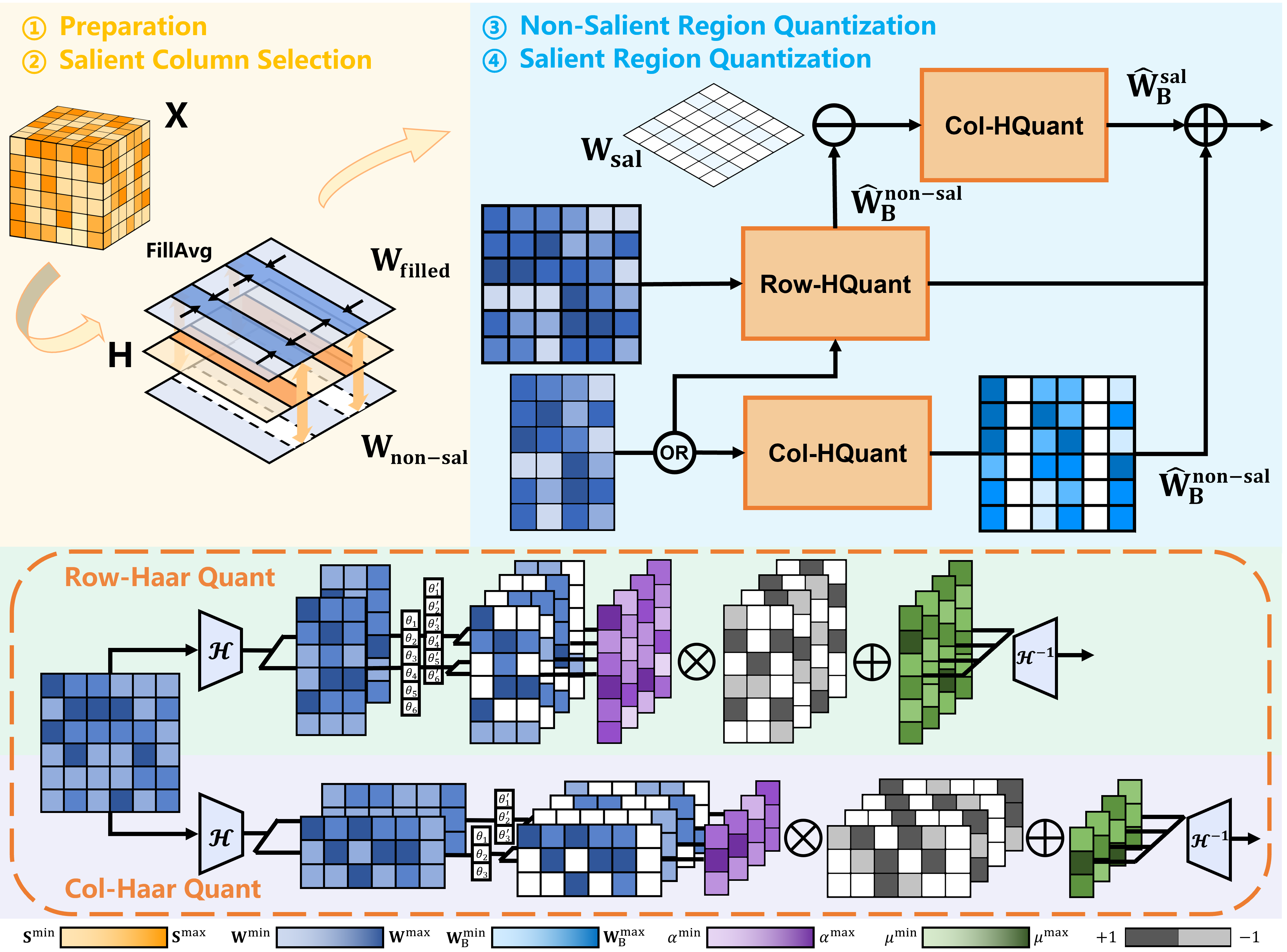}
	\caption{Overview of our HBLLM.{The HBLLM quantization process consists of four steps: preparation, salient column selection, haar transform and quantization for the non-salient part, and quantization for the salient part. Since the salient columns are excluded from the Haar transform of the non-salient part, their positions must be filled before performing row-wise Haar transforms. This is handled by a process we refer to as FillAvg, where each missing column is filled with the average of its adjacent non-salient columns. For the non-salient part, HBLLM supports flexible choice between row-wise (HBLLM-row) and column-wise (HBLLM-col) transforms. The salient part undergoes column-wise Haar transformation followed by HaarQuant for quantization.}}
	\label{fig:hbllm_overview}
\end{figure}

\subsection{Method Overview}
We define the objective of HBLLM under the binary quantization setting for LLM weights. Specifically, the quantization targets the full-precision weight matrix $\mathbf{W}_{\text{FP}} \in \mathbb{R}^{d \times d}$, where a binary diagonal mask matrix $\mathbf{M}_{\text{sal}} \in \{0,1\}^{d \times d}$ indicates which columns are selected as salient. The salient and non-salient parts are quantized in the Haar domain, and their respective quantized Haar coefficients are denoted by $\widehat{\mathbf{W}}^{\text{sal}}_{\text{B}}$ and $\widehat{\mathbf{W}}^{\text{non-sal}}_{\text{B}}$. These are then reconstructed using inverse Haar transforms $\mathcal{H}_1^{-1}$ and $\mathcal{H}_2^{-1}$.

{
The reconstruction objective of HBLLM is twofold. For the quantization of a matrix layer $\mathbf{W}$, the objective expressed in the Frobenius norm is formulated as:
\begin{equation}
	\min_{\widehat{\mathbf{W}} }\left\| \mathbf{W}\mathbf{X} - \widehat{\mathbf{W}}\mathbf{X}\right\|_{F}^2,
\end{equation}
where $\mathbf{X}$ is the input of the matrix layer. For quantization of a matrix block $\mathbf{W}_{\text{FP}}$ of  $\mathbf{W}$, the object is:}
\begin{equation}
	\min_{\mathbf{M}_{\text{sal}}, \widehat{\mathbf{W}}^{\text{sal}}_{\text{B}}, \widehat{\mathbf{W}}^{\text{non-sal}}_{\text{B}} }
	\left\| \mathbf{W}_{\text{FP}} -   \mathbf{M}_{\text{sal}}\mathcal{H}^{-1}_{1}\left(\widehat{\mathbf{W}}^{\text{sal}}_{\text{B}}\right) - (\mathbf{I}-\mathbf{M}_{\text{sal}})\mathcal{H}^{-1}_{2}\left(\widehat{\mathbf{W}}^{\text{non-sal}}_{\text{B}}\right)
	\right\|_{F}^2.
	\label{eq:hbllm_loss}
\end{equation}

When $\mathcal{H}_1 = \mathcal{H}_2$ are fixed Haar transforms, this formulation simplifies to a quantization problem entirely in the Haar domain. In this case, the objective is the same to that of BiLLM.
{ Layer-level quantization is commonly tackled with the GPTQ algorithm~\cite{OPTQ2023Frantar}. 
 }

We emphasize that our approach does not aim to solve this objective function via explicit optimization. Instead, this formulation serves as a conceptual framework that guides our method design. The actual quantization process is based on a set of heuristics and structure-aware strategies that approximate this objective in a computationally efficient and scalable manner.

\textbf{Quantization Pipeline Overview. }HBLLM integrates the Haar transform into a BiLLM-style quantization pipeline (see Algorithm~\ref{alg:hbllm_main} and Figure~\ref{fig:hbllm_overview}), consisting of the following key steps:
\begin{enumerate}[leftmargin=2em]
	\item \textbf{Preparation Phase:} Compute the column-wise importance scores using a Hessian-based saliency metric.
	\item \textbf{Salient Column Selection and Quantization(SALIENT):}
	\begin{itemize}[leftmargin=1em]
		\item Sort columns by their $\ell_2$ norm significance.
		\item Select top-${K}$ salient columns  and determine $ \mathbf{M}_{\text{sal}}$.
		\item $ \widehat{\mathbf{W}}^{\text{sal}}_{\text{B}} = \text{HaarQuant}\left(
		\mathbf{M}_{\text{sal}}\mathbf{W}
		\right)$.
		\item Choose the subset with the lowest quantization error.
	\end{itemize}
	\item \textbf{Non-Salient Region Quantization:}
	\begin{itemize}[leftmargin=1em]
		\item Fill the missing values in salient columns using adjacent averages (FillAvg).
		\item $ \widehat{\mathbf{W}}^{\text{non-sal}}_{\text{B}} = \text{HaarQuant}\left(
		\mathbf{M}_{\text{sal}}\mathbf{W}_{\text{filled}}+ \left(\mathbf{I}-\mathbf{M}_{\text{sal}}\right)\mathbf{W}
		\right)$, where $\mathbf{W}_{\text{filled}}$ is from FillAvg.
	\end{itemize}
	\item \textbf{Adjustment and Refinement:}
	\begin{itemize}[leftmargin=1em]
		\item $ \widetilde{\mathbf{W}}=\mathbf{M}_{\text{sal}} \left( \mathbf{W}- 
		\mathcal{H}^{-1}\left(  \widehat{ {\mathbf{W}}}_{B}^{\text{non-sal}} \right)
		\right).$
		\item $ \widehat{\mathbf{W}}^{\text{sal}}_{\text{B}} = \text{HaarQuant}\left(\widetilde{\mathbf{W}}\right).$
	\end{itemize}
\end{enumerate}

\begin{algorithm}[th]
	\caption{Framework of HBLLM: Details of each function are shown in Algorithm E.1}
	\label{alg:hbllm_main}
	\setstretch{1.2}
	\vspace{0.3em}
	\begin{minipage}[t]{0.56\textwidth} 
		\raggedright
		\textbf{func} HBLLM$(\mathbf{W}, \mathbf{X}, \beta, \lambda)$\\
		\textbf{Input:} $\mathbf{W} \in \mathbb{R}^{n \times m}$ - weight matrix\\
		\phantom{\textbf{Input:}} $\mathbf{X} \in \mathbb{R}^{r \times d}$ - calibration data\\
		\phantom{\textbf{Input:}} $\beta$ - block size\\
		\phantom{\textbf{Input:}} $\lambda$ - hessian regularizer\\
		\textbf{Output:} $\mathbf{B}$ - haared binarized weights
		
		\begin{algorithmic}[1]
			\State $\mathbf{H} \gets 2\mathbf{X}\mathbf{X}^\top$ \quad \text{// $\ell^2$ error hessian matrix}
			\State $\mathbf{H}^c \gets \text{Cholesky}((\mathbf{H} + \lambda \mathbf{I})^{-1})$
			\State $\mathbf{B} \gets \mathbf{0}_{n \times m}$
			\For{$b = 0, \beta, 2\beta, \dots, N$}
			\State $\mathbf{W}^b \gets \mathbf{W}_{:, b:b+\beta}$
			\State $\text{rows}\{\cdot\} \gets \text{SALIENT}(\mathbf{W}_{:, b:b+\beta}, \mathbf{H}^c)$
			\If{Row-HBLLM}
			\State $\mathbf{B}_{:, b:b+\beta} \gets \text{Row-HaarQuant}(\mathbf{W}^b, \text{rows}\{\cdot\})$
			\ElsIf{Col-HBLLM}
			\State $\mathbf{B}_{:, b:b+\beta} \gets \text{Col-HaarQuant}(\mathbf{W}^b, \text{rows}\{\cdot\})$
			\EndIf
			\State $\mathbf{E} \gets (\mathbf{W}_{:, b:b+\beta} - \mathbf{B}_{:, b:b+\beta}) / \mathbf{H}^c_{b:b+\beta, b:b+\beta}$
			\State $\mathbf{W}_{:, b+\beta:} \gets \mathbf{W}_{:, b+\beta:} - \mathbf{E} \cdot \mathbf{H}^c_{b:b+\beta, b+\beta:}$
			\EndFor
			\State \Return $\mathbf{B}$
		\end{algorithmic}
		
	\end{minipage}
	\hfill\hspace{-1em}
	\begin{minipage}[t]{0.58\textwidth} 
		\raggedright
		\textbf{func} Row-HaarQuant($\mathbf{W}, \text{rows}\{\cdot\}$)
		\begin{algorithmic}[1]
			\State $\mathbf{W}_{\mathrm{filled}} \gets \text{FillAvg}(\mathbf{W}_{:, j \notin \text{rows}},\text{rows}\{\cdot\})$
			\State $\mathbf{B}_{\text{filled}} \gets \text{HaarQuant}(\mathbf{W}_{\mathrm{filled}}, \texttt{ROW})$
			\State $\widehat{\mathbf{W}} \gets \mathbf{W} -\mathbf{B}_\text{filled}$
			\State $\mathbf{B}_{\text{salient}} \gets \text{HaarQuant}(\widehat{\mathbf{W}}_{:, j \in \text{rows}},\texttt{COL})$
			\State $\mathbf{B} \gets \mathbf{B}_{\text{salient}} + \mathbf{B}_{\text{filled}}$ 
			\State \Return $\mathbf{B}$
		\end{algorithmic}
		
		\vspace{0.05cm}
		
		\textbf{func} Col-HaarQuant$(\mathbf{W}, \text{rows}\{\cdot\})$\\
		\begin{algorithmic}[1]
			\State $\mathbf{B}_{\text{unsalient}} \gets \text{HaarQuant}(\mathbf{W}_{:, j \notin \text{rows}}, \texttt{COL})$
			\State $\widehat{\mathbf{W}} \gets \mathbf{W} -\mathbf{B}_\text{filled}$
			\State $\mathbf{B}_{\text{salient}} \gets \text{HaarQuant}(\widehat{\mathbf{W}}_{:, j \in \text{rows}},\texttt{COL})$
			\State $\mathbf{B} \gets \mathbf{B}_{\text{salient}} + \mathbf{B}_{\text{unsalient}}$ 
			\State \Return $\mathbf{B}$
		\end{algorithmic}
	\end{minipage}
	\label{algo:hbllm_framework}
\end{algorithm}

\subsection{HaarQuant: One-Bit Quantization in the Wavelet Domain}

To boost expressiveness, we apply Haar wavelet transform to the weight matrix of linear layers, generating a frequency-domain coefficient matrix, followed by group-wise $1$-bit quantization.

To address limited numerical expressiveness, HBLLM introduces the HaarQuant algorithm. HaarQuant consists of three stages.

\textbf{Haar Transform. }A row of weights $\mathbf{W}$ is decomposed into low- and high-frequency coefficients via 1D Haar transform $\mathcal{H}$:
\begin{equation}
	\widehat{\mathbf{W}} = \mathcal{H}\left({\mathbf{W}} \right)
	= \left[ 
	\mathcal{H}_{\text{low-pass}}\left({\mathbf{W}} \right),
	\mathcal{H}_{\text{high-pass}}\left({\mathbf{W}} \right)
	\right],
	\label{eq:haar_transform}
\end{equation}

where $\widehat{\mathbf{W}}$ is the Haar coefficient of $\mathbf{W}$, 
$\mathcal{H}_{\text{low-pass}}\left({\mathbf{W}} \right)$ and $\mathcal{H}_{\text{high-pass}}\left({\mathbf{W}} \right)$ are  low- and high-frequency coefficients, respecively.

\textbf{Frequency-Aware Multi-Parameter Intra-Row Grouping. }
For each row, boundary candidates determined by the row are enumerated, and the best grouping with minimal quantization error is selected. Furthermore, we split the rows by frequency bands. This adaptive strategy captures intra-row structural differences better than global uniform boundaries used in BiLLM.

\textbf{Coefficient Quantization. }Each group $\widehat{\mathbf{W}}_\mathrm{FP}$ is quantized using sign-based binarization centered on its mean:
\begin{equation}
	\widehat{\mathbf{W}}_\mathrm{B} = \alpha \cdot \operatorname{sign}({\widehat{\mathbf{W}}_\mathrm{FP}} - \mu),
	\label{eq:haar-binarization}
\end{equation}
where $\alpha \in \mathbb{R}^{d}$ is the row-wise scaling factor and 
$\mu$ is the group-wise mean and $\widehat{\mathbf{W}}_\mathrm{B}$ is the result.

\subsection{Structure-aware Grouping Strategies }
To enhance the fidelity and adaptability of binary quantization under structural constraints, HBLLM introduces two structure-aware grouping strategies that operate along both column and row dimensions of the weight matrix.

\textbf{Saliency-Driven Column Selection via $\ell_2$ Norm.} This strategy is used during salient column identification and quantization to overcome the limitations of prior heuristics based on fixed thresholds or simple magnitude criteria.
\begin{itemize}[leftmargin=1em]
	\item Columns are ranked by their $\ell_2$-norm scores, which correlate with their overall contribution to activation magnitude.
	\item The top-$K$ columns are selected as salient and quantized in the Haar-transformed domain using column-wise transforms.
\end{itemize}
This approach helps preserve activation-critical directions, especially those dominated by outlier weights.

\textbf{Frequency-Aware Multi-Parameter Intra-Row Grouping.} This strategy is used during Haar domain quantization, where conventional row grouping lacks sensitivity to structural variations in weight distributions.
\begin{itemize}[leftmargin=1em]
	\item Each row is first decomposed into high- and low-frequency components based on Haar subbands.
	\item Within each frequency band, coefficients are adaptively split into dense and sparse groups using band-specific, data-driven thresholds.
\end{itemize}
This grouping effectively doubles the number of quantization subgroups per row, enabling finer granularity and better error control.

Together, these strategies facilitate fine-grained, structure-preserving quantization across both dimensions of the weight matrix. To further guide saliency-based partitioning, we adopt the parameter importance metric used in BiLLM, defined as:
$s_i={w_i^2}/{[\mathbf{H}^{-1}]^2_{ii}}$,
where $\mathbf{H}$ denotes the Hessian matrix of the layer, $w_i$ is the full-precision value of the $i$-th parameter, and $[\mathbf{H}^{-1}]_{ii}$ is the $i$-th diagonal entry of the inverse Hessian.

This metric reflects the relative sensitivity of the loss to changes in each parameter: higher values indicate greater influence on the model's output, and thus prioritize that weight for accurate reconstruction.

\subsection{Intra-frequency-band Mean Sharing  }
To reduce storage overhead, HBLLM shares a single mean value among $2$ groups in the same frequency band within each row:
$\mu_{\text{shared}} = \frac{1}{n_1 + n_2} \left( \sum_{i=1}^{n_1} x_i + \sum_{j=1}^{n_2} y_j \right).$ It not only reduces per-parameter storage by $0.25$ bits, but also maintains accuracy even slightly improving downstream task performance. This optimization achieves a trade-off between compression and accuracy, improving deployment viability.

\subsection{Efficient Haar Implementation via Local Convolutions }
Instead of costly matrix multiplication, HBLLM implements Haar transform using fixed local convolutions. There are
only two predefined 1D kernels, $[1/2, 1/2]$ and $[1/2, -1/2]$, whose 
kernel size is $2$. Furthermore, it can be hardcoded into the model for zero runtime initialization and 
no training or storage is needed. 
In complexity comparison, HBLLM needs $\mathcal{O}(d)$ operatons via convolutional sliding window, while 
FrameQuant needs $\mathcal{O}(d^2)$ operations. As a result, HBLLM  significantly lowers inference cost and is ideal for edge deployment.

\begin{table}
	\caption{Comparison of perplexity and average accuracy across models and methods}
	\vspace{0.2em}
	\label{tab:summary-table}
	\scriptsize  
	\begin{minipage}{0.48\textwidth}
		\setlength{\tabcolsep}{3pt}
		\begin{tabular}{c c c !{\vrule} c c c !{\vrule} c}
			\toprule
			\multicolumn{3}{c}{\textbf{LLaMA1}} & 
			\multicolumn{3}{c}{\textbf{Perplexity↓}} & 
			\multirow{2}{*}{\textbf{AvgQA↑}} \\
			\cmidrule(lr){1-3}
			\cmidrule(lr){4-6}
			\textbf{Size} & \textbf{Method} & \multicolumn{1}{c}{\textbf{W-bits}} & \textbf{C4} & \textbf{Wiki2} & \multicolumn{1}{c}{\textbf{PTB}} & \\
			\midrule
			\multirow{8}{*}{7B}
			& FullPrecision       & 16.00 & 6.71   & 5.68   & 35.80   & 65.62 \\
			\cmidrule(l{2pt}r{2pt}){2-7}
			& FrameQuant          & 2.20  & 10.89   & 9.96    & 104.7   & 56.19 \\
			& PB-LLM              & 1.70  & 90.19   & 113.4   & 830.0   & 35.71 \\
			& BiLLM               & 1.09  & 43.74   & 44.85   & 369.3   & 40.01 \\
			& ARB-LLM\textsubscript{X}  & 1.09  & 22.80   & 24.70   & 240.5   & 45.65 \\
			& ARB-LLM\textsubscript{RC} & 1.09  & 15.13   & 13.45   & 155.8   & 52.23 \\
			& \cellcolor{gray!15}\hspace{-\tabcolsep} HBLLM-row & \cellcolor{gray!15}\hspace{-\tabcolsep} 1.09 & \cellcolor{gray!15}\hspace{-\tabcolsep} \textbf{9.49} & \cellcolor{gray!15}\hspace{-\tabcolsep} \textbf{8.82} & \cellcolor{gray!15}\hspace{-\tabcolsep} \textbf{88.86} & \cellcolor{gray!15}\hspace{-\tabcolsep} \textbf{57.48}\\
			& \cellcolor{gray!15}\hspace{-\tabcolsep} HBLLM-col & \cellcolor{gray!15}\hspace{-\tabcolsep} 1.00 & \cellcolor{gray!15}\hspace{-\tabcolsep} 10.38 & \cellcolor{gray!15}\hspace{-\tabcolsep} 9.67 & \cellcolor{gray!15}\hspace{-\tabcolsep} 117.7 & \cellcolor{gray!15}\hspace{-\tabcolsep} 54.03 \\
			\midrule
			
			\multirow{8}{*}{13B}
			& FullPrecision       & 16.00 & 6.24   & 5.09   & 25.36   & 68.09  \\ 
			\cmidrule(l{2pt}r{2pt}){2-7}
			& FrameQuant          & 2.20  & 8.79   & 7.84   & 50.69   &  60.69 \\ 
			& PB-LLM              & 1.70  & 38.41  & 46.02  & 190.2   &  40.39 \\ 
			& BiLLM               & 1.10  & 13.93  & 14.99  & 69.75   &  50.89 \\ 
			& ARB-LLM\textsubscript{X}  & 1.10  & N/A  & N/A  &  N/A  & N/A \\ 
			& ARB-LLM\textsubscript{RC} & 1.10  & 10.68  & 10.19  & 43.85   &  59.58\\ 
			& \cellcolor{gray!15}\hspace{-\tabcolsep} HBLLM-row & \cellcolor{gray!15}\hspace{-\tabcolsep} 1.09 & \cellcolor{gray!15}\hspace{-\tabcolsep} \textbf{7.62} & \cellcolor{gray!15}\hspace{-\tabcolsep} \textbf{6.68} & \cellcolor{gray!15}\hspace{-\tabcolsep} \textbf{34.94} & \cellcolor{gray!15}\hspace{-\tabcolsep} \textbf{62.57}\\ 
			& \cellcolor{gray!15}\hspace{-\tabcolsep} HBLLM-col & \cellcolor{gray!15}\hspace{-\tabcolsep} 1.00 & \cellcolor{gray!15}\hspace{-\tabcolsep} 7.77 & \cellcolor{gray!15}\hspace{-\tabcolsep} 6.98 & \cellcolor{gray!15}\hspace{-\tabcolsep} 37.62 & \cellcolor{gray!15}\hspace{-\tabcolsep} 61.25 \\ 
			\midrule
			
			\multirow{8}{*}{30B}
			& FullPrecision       & 16.00 & 5.62   & 4.10   & 21.35   &  71.06\\ 
			\cmidrule(l{2pt}r{2pt}){2-7}
			& FrameQuant          & 2.20  &   7.35   &  6.32  &  28.69  &  65.13\\ 
			& PB-LLM              & 1.70  & 21.73  & 25.87  & 127.1   & 47.22 \\ 
			& BiLLM               & 1.11  & 10.27  & 10.55  & 41.76   & 58.07 \\ 
			& ARB-LLM\textsubscript{X}  & 1.11  &  N/A  & N/A  &  N/A  & N/A  \\ 
			& ARB-LLM\textsubscript{RC} & 1.11  & 8.49   & 7.79   &  30.98  & 64.49 \\ 
			& \cellcolor{gray!15}\hspace{-\tabcolsep} HBLLM-row & \cellcolor{gray!15}\hspace{-\tabcolsep} 1.10 & \cellcolor{gray!15}\hspace{-\tabcolsep} \textbf{6.88} & \cellcolor{gray!15}\hspace{-\tabcolsep} \textbf{5.82} & \cellcolor{gray!15}\hspace{-\tabcolsep} \textbf{25.95} & \cellcolor{gray!15}\hspace{-\tabcolsep} \textbf{66.76} \\ 
			& \cellcolor{gray!15}\hspace{-\tabcolsep} HBLLM-col & \cellcolor{gray!15}\hspace{-\tabcolsep} 1.00 & \cellcolor{gray!15}\hspace{-\tabcolsep} 7.03 & \cellcolor{gray!15}\hspace{-\tabcolsep} 6.03 & \cellcolor{gray!15}\hspace{-\tabcolsep} 26.65 & \cellcolor{gray!15}\hspace{-\tabcolsep} 64.86 \\ 
			\midrule
			
			\multirow{8}{*}{65B}
			& FullPrecision       & 16.00 & 5.31  & 3.53  & 21.11  & 72.27 \\
			\cmidrule(l{2pt}r{2pt}){2-7}
			& FrameQuant          & 2.20  & 6.69  & 5.55  & 27.48  & 68.58 \\
			& PB-LLM              & 1.70  & 12.66 & 12.76 & 99.67  & 62.48 \\
			& BiLLM               & 1.10  & 9.26  & 8.58  & 41.93  & 62.05 \\
			& ARB-LLM\textsubscript{X}  & 1.10   & N/A   & N/A   & N/A    & N/A  \\
			& ARB-LLM\textsubscript{RC} & 1.10  & 7.48  & 6.47  & 29.14  & 68.53 \\
			& \cellcolor{gray!15}\hspace{-\tabcolsep} HBLLM-row & \cellcolor{gray!15}\hspace{-\tabcolsep} 1.09 & \cellcolor{gray!15}\hspace{-\tabcolsep} \textbf{6.28} & \cellcolor{gray!15}\hspace{-\tabcolsep} \textbf{5.07} & \cellcolor{gray!15}\hspace{-\tabcolsep} \textbf{24.11} & \cellcolor{gray!15}\hspace{-\tabcolsep} \textbf{69.18} \\
			& \cellcolor{gray!15}\hspace{-\tabcolsep} HBLLM-col & \cellcolor{gray!15}\hspace{-\tabcolsep} 1.00 & \cellcolor{gray!15}\hspace{-\tabcolsep} 6.44 & \cellcolor{gray!15}\hspace{-\tabcolsep} 5.26 & \cellcolor{gray!15}\hspace{-\tabcolsep} 30.38 & \cellcolor{gray!15}\hspace{-\tabcolsep} 67.83 \\
			\bottomrule
		\end{tabular}
		
		\begin{tabular}{c c c !{\vrule} c c c !{\vrule} c}
			\toprule
			\multicolumn{3}{c}{\textbf{LLaMA2}} & 
			\multicolumn{3}{c}{\textbf{Perplexity↓}} & 
			\multirow{2}{*}{\textbf{AvgQA↑}} \\
			\cmidrule(lr){1-3}
			\cmidrule(lr){4-6}
			\textbf{Size} & \textbf{Method} & \multicolumn{1}{c}{\textbf{W-bits}} & \textbf{C4} & \textbf{Wiki2} & \multicolumn{1}{c}{\textbf{PTB}} & \\
			\midrule
			\multirow{8}{*}{7B}
			& FullPrecision       & 16.00 & 8.66   & 6.94   & 37.86   & 65.54 \\ 
			\cmidrule(l{2pt}r{2pt}){2-7}
			& FrameQuant          & 2.20  & 14.66  & 13.34  & 177.1   & 52.75 \\ 
			& PB-LLM              & 1.70  & 63.95  & 55.40  & 486.2   & 36.54 \\ 
			& BiLLM               & 1.08  & 33.97  & 31.38  & 373.0   & 42.11 \\ 
			& ARB-LLM\textsubscript{X}  & 1.08  & 26.55  & 21.74  & 314.2   & 45.41 \\ 
			& ARB-LLM\textsubscript{RC} & 1.08  & 17.87  & 15.85  & 462.2   & 46.71 \\ 
			& \cellcolor{gray!15}\hspace{-\tabcolsep} HBLLM-row & \cellcolor{gray!15}\hspace{-\tabcolsep} 1.07 & \cellcolor{gray!15}\hspace{-\tabcolsep} \textbf{11.75} & \cellcolor{gray!15}\hspace{-\tabcolsep} \textbf{10.52} & \cellcolor{gray!15}\hspace{-\tabcolsep} \textbf{89.23} & \cellcolor{gray!15}\hspace{-\tabcolsep} \textbf{57.74} \\ 
			& \cellcolor{gray!15}\hspace{-\tabcolsep} HBLLM-col & \cellcolor{gray!15}\hspace{-\tabcolsep} 1.00 & \cellcolor{gray!15}\hspace{-\tabcolsep} 12.51 & \cellcolor{gray!15}\hspace{-\tabcolsep} 11.33 & \cellcolor{gray!15}\hspace{-\tabcolsep} 150.6 & \cellcolor{gray!15}\hspace{-\tabcolsep} 54.09 \\ 
			\midrule

			\multirow{8}{*}{13B}
			& FullPrecision       & 16.00 & 6.18   & 4.88   & 43.02   & 69.18 \\ 
			\cmidrule(l{2pt}r{2pt}){2-7}
			& FrameQuant          & 2.20  & 9.40   & 7.80   & 109.3   &  61.35\\ 
			& PB-LLM              & 1.70  & 313.4  & 289.4  & 934.4   &  32.91\\ 
			& BiLLM               & 1.08  & 22.17  & 19.57  & 303.4   &  46.76\\ 
			& ARB-LLM\textsubscript{X}  & 1.08  & N/A  & N/A  &  N/A  & N/A  \\ 
			& ARB-LLM\textsubscript{RC} & 1.08  & 11.90  & 10.98  & 151.8   & 57.35 \\ 
			& \cellcolor{gray!15}\hspace{-\tabcolsep} HBLLM-row & \cellcolor{gray!15}\hspace{-\tabcolsep} 1.07 & \cellcolor{gray!15}\hspace{-\tabcolsep} \textbf{7.82} & \cellcolor{gray!15}\hspace{-\tabcolsep} \textbf{6.71} & \cellcolor{gray!15}\hspace{-\tabcolsep} \textbf{61.75} & \cellcolor{gray!15}\hspace{-\tabcolsep} \textbf{63.61} \\ 
			& \cellcolor{gray!15}\hspace{-\tabcolsep} HBLLM-col & \cellcolor{gray!15}\hspace{-\tabcolsep} 1.00 & \cellcolor{gray!15}\hspace{-\tabcolsep} 8.28 & \cellcolor{gray!15}\hspace{-\tabcolsep} 7.00 & \cellcolor{gray!15}\hspace{-\tabcolsep} 69.74 & \cellcolor{gray!15}\hspace{-\tabcolsep} 62.04\\
			\midrule
			
			\multirow{8}{*}{70B}
			& FullPrecision       & 16.00 & 5.24  & 3.32  & 21.49  & 72.96 \\
			\cmidrule(l{2pt}r{2pt}){2-7}
			& FrameQuant          & 2.20  & N/A    & N/A    & N/A     & N/A \\
			& PB-LLM              & 1.70  & N/A   & N/A   & N/A  & 54.26 \\
			& BiLLM               & 1.09  & 15.57 & 15.86 & 71.03  & 55.81 \\
			& ARB-LLM\textsubscript{X}  & 1.09  & N/A   & N/A   & N/A    & N/A   \\
			& ARB-LLM\textsubscript{RC} & 1.09  & 7.26  & 6.00  & 28.43  & 68.77 \\
			& \cellcolor{gray!15}\hspace{-\tabcolsep} HBLLM-row & \cellcolor{gray!15}\hspace{-\tabcolsep} 1.08 & \cellcolor{gray!15}\hspace{-\tabcolsep} \textbf{6.18} & \cellcolor{gray!15}\hspace{-\tabcolsep} \textbf{4.82} & \cellcolor{gray!15}\hspace{-\tabcolsep} \textbf{24.69} & \cellcolor{gray!15}\hspace{-\tabcolsep} \textbf{70.01} \\
			& \cellcolor{gray!15}\hspace{-\tabcolsep} HBLLM-col & \cellcolor{gray!15}\hspace{-\tabcolsep} 1.00 & \cellcolor{gray!15}\hspace{-\tabcolsep} 6.63 & \cellcolor{gray!15}\hspace{-\tabcolsep} 5.04 & \cellcolor{gray!15}\hspace{-\tabcolsep} 26.31 & \cellcolor{gray!15}\hspace{-\tabcolsep} 68.61 \\ 
			
			\bottomrule
		\end{tabular}	
		
	\end{minipage}
	\hspace{0.02\textwidth}
	\begin{minipage}{0.48\textwidth}
		\setlength{\tabcolsep}{3pt}
		\begin{tabular}{c c c !{\vrule} c c c !{\vrule} c}
			\toprule
			\multicolumn{3}{c}{\textbf{LLaMA3}} & 
			\multicolumn{3}{c}{\textbf{Perplexity↓}} & 
			\multirow{2}{*}{\textbf{AvgQA↑}} \\
			\cmidrule(lr){1-3}
			\cmidrule(lr){4-6}
			\textbf{Size} & \textbf{Method} & \multicolumn{1}{c}{\textbf{W-bits}} & \textbf{C4} & \textbf{Wiki2} & \multicolumn{1}{c}{\textbf{PTB}} & \\
			\midrule
			\multirow{8}{*}{8B}
			& FullPrecision       & 16.00 & 11.90 & 8.29 & 13.07 & 68.94 \\
			\cmidrule(l{2pt}r{2pt}){2-7}
			& FrameQuant          & 2.20  & 28.44 & 23.36 & 40.33 & 52.27 \\
			& PB-LLM              & 1.70  & 111.7 & 141.5 & 171.1 & 36.83 \\
			& BiLLM               & 1.06  & 53.67 & 56.24 & 81.27 & 41.84 \\
			& ARB-LLM\textsubscript{X}  & 1.06  & 48.45 & 37.90 & 52.59 & 43.40 \\
			& ARB-LLM\textsubscript{RC} & 1.06  & 34.44 & 30.24 & 45.23 & 49.08 \\
			& \cellcolor{gray!15}\hspace{-\tabcolsep} HBLLM-row & \cellcolor{gray!15}\hspace{-\tabcolsep} 1.06 & \cellcolor{gray!15}\hspace{-\tabcolsep} \textbf{20.09} & \cellcolor{gray!15}\hspace{-\tabcolsep} \textbf{16.18} & \cellcolor{gray!15}\hspace{-\tabcolsep} \textbf{22.83} & \cellcolor{gray!15}\hspace{-\tabcolsep} \textbf{54.80} \\
			& \cellcolor{gray!15}\hspace{-\tabcolsep} HBLLM-col & \cellcolor{gray!15}\hspace{-\tabcolsep} 1.00 & \cellcolor{gray!15}\hspace{-\tabcolsep} 22.18 & \cellcolor{gray!15}\hspace{-\tabcolsep} 17.80 & \cellcolor{gray!15}\hspace{-\tabcolsep} 26.38 & \cellcolor{gray!15}\hspace{-\tabcolsep} 51.43 \\
			\midrule
			
			\multirow{8}{*}{70B}
			& FullPrecision       & 16.00 & 6.61   & 2.85   & 7.74    & 74.62 \\
			\cmidrule(l{2pt}r{2pt}){2-7}
			& FrameQuant          & 2.20  & N/A    & N/A    & N/A     & N/A \\
			& PB-LLM              & 1.70  & 33.56  & 28.93  & 44.38   & 47.45 \\
			& BiLLM               & 1.09  & 385.8 & 137.6 & 129.5  & 34.18 \\
			& ARB-LLM\textsubscript{X} & 1.09  & N/A    & N/A    & N/A     & N/A \\
			& ARB-LLM\textsubscript{RC}  & 1.09  & 12.80  & 10.24  & 12.76   & \textbf{63.90} \\
			& \cellcolor{gray!15}\hspace{-\tabcolsep} HBLLM-row & \cellcolor{gray!15}\hspace{-\tabcolsep} 1.08 & \cellcolor{gray!15}\hspace{-\tabcolsep} \textbf{10.87} & \cellcolor{gray!15}\hspace{-\tabcolsep} \textbf{8.08} & \cellcolor{gray!15}\hspace{-\tabcolsep} \textbf{11.44} & \cellcolor{gray!15}\hspace{-\tabcolsep} 56.45 \\
			& \cellcolor{gray!15}\hspace{-\tabcolsep} HBLLM-col & \cellcolor{gray!15}\hspace{-\tabcolsep} 1.00 & \cellcolor{gray!15}\hspace{-\tabcolsep} 13.69 & \cellcolor{gray!15}\hspace{-\tabcolsep} 9.09 & \cellcolor{gray!15}\hspace{-\tabcolsep} 14.26 & \cellcolor{gray!15}\hspace{-\tabcolsep} 55.89 \\
			\bottomrule
		\end{tabular}
		
		\begin{tabular}{c c c !{\vrule} c c c !{\vrule} c}
			\toprule
			\multicolumn{3}{c}{\textbf{OPT}} & 
			\multicolumn{3}{c}{\textbf{Perplexity↓}} & 
			\multirow{2}{*}{\textbf{AvgQA↑}} \\
			\cmidrule(lr){1-3}
			\cmidrule(lr){4-6}
			\textbf{Size} & \textbf{Method} & \multicolumn{1}{c}{\textbf{W-bits}} & \textbf{C4} & \textbf{Wiki2} & \multicolumn{1}{c}{\textbf{PTB}} & \\
			\midrule
			\multirow{8}{*}{1.3B}
			& FullPrecision       & 16.00  & 13.45   & 14.62   & 16.41   & 52.54 \\
			\cmidrule(l{2pt}r{2pt}){2-7}
			& FrameQuant          & 2.20  & 24.29   & 27.15   & 30.45   & 44.48 \\
			& PB-LLM              & 1.70  & 186.9   & 309.0   & 286.3   & 33.44 \\
			& BiLLM               & 1.09  & 56.24   & 68.43   & 119.2   & 38.39 \\
			& ARB-LLM\textsubscript{X}  & 1.09  & 43.23   & 53.55   & 67.96   & 41.42 \\
			& ARB-LLM\textsubscript{RC} & 1.09  & 24.23   & 28.77   & 33.32   & 45.28 \\
			& \cellcolor{gray!15}\hspace{-\tabcolsep} HBLLM-row & \cellcolor{gray!15}\hspace{-\tabcolsep} 1.07 & \cellcolor{gray!15}\hspace{-\tabcolsep} \textbf{19.30} & \cellcolor{gray!15}\hspace{-\tabcolsep} \textbf{21.68} & \cellcolor{gray!15}\hspace{-\tabcolsep} \textbf{25.34} & \cellcolor{gray!15}\hspace{-\tabcolsep} \textbf{46.35} \\
			& \cellcolor{gray!15}\hspace{-\tabcolsep} HBLLM-col & \cellcolor{gray!15}\hspace{-\tabcolsep} 1.00 & \cellcolor{gray!15}\hspace{-\tabcolsep} 21.92 & \cellcolor{gray!15}\hspace{-\tabcolsep} 24.08 & \cellcolor{gray!15}\hspace{-\tabcolsep} 27.28 & \cellcolor{gray!15}\hspace{-\tabcolsep} 44.70 \\
			\midrule
			
			\multirow{8}{*}{2.7B}
			& FullPrecision       & 16.00 & 12.06   & 12.47   & 14.61   & 54.95 \\
			\cmidrule(l{2pt}r{2pt}){2-7}
			& FrameQuant          & 2.20  & 17.86   & 18.24   & 22.60   & \textbf{49.58} \\
			& PB-LLM              & 1.70  & 165.1   & 216.8   & 160.4   & 37.62 \\
			& BiLLM               & 1.10  & 42.92   & 55.75   & 103.2   & 40.02 \\
			& ARB-LLM\textsubscript{X}  & 1.10  & 30.02   & 34.15   & 41.35   & 44.60 \\
			& ARB-LLM\textsubscript{RC} & 1.10  & 18.02   & 19.53   & 24.46   & 49.53 \\
			& \cellcolor{gray!15}\hspace{-\tabcolsep} HBLLM-row & \cellcolor{gray!15}\hspace{-\tabcolsep} 1.09 & \cellcolor{gray!15}\hspace{-\tabcolsep} \textbf{15.70} & \cellcolor{gray!15}\hspace{-\tabcolsep} \textbf{16.85} & \cellcolor{gray!15}\hspace{-\tabcolsep} \textbf{19.54} & \cellcolor{gray!15}\hspace{-\tabcolsep} 48.80 \\
			& \cellcolor{gray!15}\hspace{-\tabcolsep} HBLLM-col & \cellcolor{gray!15}\hspace{-\tabcolsep} 1.00 & \cellcolor{gray!15}\hspace{-\tabcolsep} 17.28 & \cellcolor{gray!15}\hspace{-\tabcolsep} 18.80 & \cellcolor{gray!15}\hspace{-\tabcolsep} 22.63 & \cellcolor{gray!15}\hspace{-\tabcolsep} 48.56 \\
			\midrule
			
			\multirow{8}{*}{6.7B}
			& FullPrecision       & 16.00 & 10.68  & 10.86   & 12.73   & 58.95 \\
			\cmidrule(l{2pt}r{2pt}){2-7}
			& FrameQuant          & 2.20  & 14.53   & 14.59   & 18.71   & 53.77 \\
			& PB-LLM              & 1.70  & 122.9   & 206.7   & 222.3   & 34.87 \\
			& BiLLM               & 1.11  &  39.96  & 54.91   & 90.10   & 37.40 \\
			& ARB-LLM\textsubscript{X}  & 1.11  & 19.39   & 19.50   & 24.78   & 49.79 \\
			& ARB-LLM\textsubscript{RC} & 1.11  & 14.29   & 15.16   & 17.92   & 53.76 \\
			& \cellcolor{gray!15}\hspace{-\tabcolsep} HBLLM-row & \cellcolor{gray!15}\hspace{-\tabcolsep} 1.10 & \cellcolor{gray!15}\hspace{-\tabcolsep} \textbf{12.56} & \cellcolor{gray!15}\hspace{-\tabcolsep} \textbf{13.04} & \cellcolor{gray!15}\hspace{-\tabcolsep} \textbf{15.26} & \cellcolor{gray!15}\hspace{-\tabcolsep} \textbf{56.17} \\
			& \cellcolor{gray!15}\hspace{-\tabcolsep} HBLLM-col & \cellcolor{gray!15}\hspace{-\tabcolsep} 1.00 & \cellcolor{gray!15}\hspace{-\tabcolsep} 13.29 & \cellcolor{gray!15}\hspace{-\tabcolsep} 13.67 & \cellcolor{gray!15}\hspace{-\tabcolsep} 15.70 & \cellcolor{gray!15}\hspace{-\tabcolsep} 54.44 \\
			\midrule
			
			\multirow{8}{*}{13B}
			& FullPrecision       & 16.00 &  10.16  &  10.13 & 11.89  & 58.41 \\
			\cmidrule(l{2pt}r{2pt}){2-7}
			& FrameQuant          & 2.20  & 12.26   & 12.51   & 14.59   & 55.42 \\
			& PB-LLM              & 1.70  & 42.89  & 81.02  & 94.98  & 39.50 \\
			& BiLLM               & 1.13  & 17.01  & 18.34  & 21.56  & 49.82 \\
			& ARB-LLM\textsubscript{X}  & 1.13  & N/A  & N/A  &  N/A  & N/A  \\
			& ARB-LLM\textsubscript{RC} & 1.13 & 12.60  & 13.14 & 15.14  & 55.35 \\
			& \cellcolor{gray!15}\hspace{-\tabcolsep} HBLLM-row & \cellcolor{gray!15}\hspace{-\tabcolsep} 1.12 & \cellcolor{gray!15}\hspace{-\tabcolsep} \textbf{11.47} & \cellcolor{gray!15}\hspace{-\tabcolsep} \textbf{11.72} & \cellcolor{gray!15}\hspace{-\tabcolsep} \textbf{13.78} & \cellcolor{gray!15}\hspace{-\tabcolsep} \textbf{55.91} \\
			& \cellcolor{gray!15}\hspace{-\tabcolsep} HBLLM-col & \cellcolor{gray!15}\hspace{-\tabcolsep} 1.00 & \cellcolor{gray!15}\hspace{-\tabcolsep} 11.71 & \cellcolor{gray!15}\hspace{-\tabcolsep} 12.34 & \cellcolor{gray!15}\hspace{-\tabcolsep} 14.13 & \cellcolor{gray!15}\hspace{-\tabcolsep} 55.66 \\
			\midrule
			\multirow{8}{*}{30B}
			& FullPrecision       & 16.00 & 9.60 & 9.56 & 11.50 & 62.09 \\
			\cmidrule(l{2pt}r{2pt}){2-7}
			& FrameQuant          & 2.20  & 10.92 & 11.15 & 13.25 & 59.62 \\
			& PB-LLM              & 1.70  & 21.60 & 28.62 & 45.63 & 46.14 \\
			& BiLLM               & 1.06  & 13.43 & 13.44 & 16.66 &  54.22\\
			& ARB-LLM\textsubscript{X}  & 1.06 & N/A  & N/A  &  N/A  & N/A  \\
			& ARB-LLM\textsubscript{RC} & 1.06  & 11.18 & 10.94 & 13.27 & 58.59 \\
			& \cellcolor{gray!15}\hspace{-\tabcolsep} HBLLM-row & \cellcolor{gray!15}\hspace{-\tabcolsep} 1.06 & \cellcolor{gray!15}\hspace{-\tabcolsep} \textbf{10.41} & \cellcolor{gray!15}\hspace{-\tabcolsep} \textbf{10.13} & \cellcolor{gray!15}\hspace{-\tabcolsep} \textbf{12.58} & \cellcolor{gray!15}\hspace{-\tabcolsep} \textbf{60.04} \\
			& \cellcolor{gray!15}\hspace{-\tabcolsep} HBLLM-col & \cellcolor{gray!15}\hspace{-\tabcolsep} 1.00 & \cellcolor{gray!15}\hspace{-\tabcolsep} 10.53 & \cellcolor{gray!15}\hspace{-\tabcolsep} 10.29 & \cellcolor{gray!15}\hspace{-\tabcolsep} 12.75 & \cellcolor{gray!15}\hspace{-\tabcolsep} 58.91 \\
			\bottomrule
		\end{tabular}

	\end{minipage}
	
	\begin{minipage}{0.98\linewidth}
		\vspace{2pt}
		\footnotesize
		\textit{Note:} All methods are calibrated on C4 with $128$ samples and a sequence length of $2048$. A block size of $128$ is used for channel-wise quantization, as commonly done in prior work. N/A: ARB-LLM\textsubscript{X} method cannot run on a single 3090 GPU - 24GB. {W-bits is the average  weight overhead per weight. For more details, please refer to the appendix D. }
	\end{minipage}
\end{table}
\vspace{-1em}

\section{Experiments}
\subsection{Experimental Settings}
\textbf{Models and Evaluation Datasets.} In our study, we evaluate HBLLM on various models, including those from the OPT, LLaMA-$1$, LLaMA-$2$, and LLaMA-$3$, as well as the recently introduced f-R1-Distill-Llama-$8$B. Specifically, we utilize the OPT models with $1.3$B and $2.7$B parameters, the LLaMA-$1$ and LLaMA-$2$ models with $7$B and $13$B parameters for our evaluations, and the LLaMA-$3$ model with $8$B parameters. We measure language modeling capabilities of these models by evaluating their perplexity on the C$4$\cite{c42020Raffel}, WikiText$2$\cite{wiki22017Merity} and PTB\cite{ptb1994Marcus} datasets. Additionally, we assess zero-shot accuracy on various Common Sense Reasoning Tasks such as PIQA\cite{Piqa2020Bisk}, BoolQ\cite{Boolq2019Clark}, OpenBookQA\cite{obqa2018Mihaylov}, WinoGrande\cite{Winogrande2021Sakaguchi}, ARC-e, ARC-c\cite{arc2018Clark}, HellaSwag\cite{HellaSwag2019Zellers}, which are commonly used for evaluating the performance of LLM quantization methods. To further enhance evaluation coverage, we also include COPA\cite{COPA2011Roemmele} for causal reasoning and LAMBADA\cite{LAMBADA2016Paperno} for long-context language modeling. All evaluations are conducted using the open-source LLM evaluation framework, LM-Evaluation-Harness\cite{Fewshot2021Mukherjee}.

\textbf{Details of Experiments.} All experiments are conducted with PyTorch on NVIDIA GeForce RTX $3090$ GPUs with $24$GB of memory. For the calibration data, we follow the settings adopted in GPTQ and BiLLM, selecting $128$ samples from the C$4$ dataset, with a sequence length of $2048$. During quantization, we set the block size to $128$ in BiLLM, PB-LLM, ARB-LLM, and HBLLM.

\textbf{Baselines.} We compare HBLLM against several state-of-the-art LLM binarization methods, including BiLLM, ARB-LLM and PB-LLM, ensuring that all implementations adhere to the details provided in their respective papers. BiLLM, ARB-LLM and PB-LLM all utilize the PTQ approach for model calibration through OBQ based method of GPTQ. For ARB-LLM, we evaluate two of its best-performing variants, ARB-LLM\textsubscript{X} and ARB-LLM\textsubscript{RC}. {{Both ARB-LLM\textsubscript{x} and ARB-LLM\textsubscript{RC} employ the salient column bitmap and group bitmap (CGB) for better performance.} For PB-LLM, which allows variable ratios of salient weights to enhance accuracy, we have set the ratio of salient weights to $10\%$ to ensure the average bit width of weight parameters remains below $2$ bits. Given the significant accuracy improvements demonstrated by HBLLM over traditional binarization techniques, we also include a comparison with a leading method using orthogonal transforms: FrameQuant. For FrameQuant, quantization is performed not in the original weight space but in the structured orthogonal basis constructed through Fusion Frames. We evaluate two configurations: FrameQuant $(r=1.0)$ and FrameQuant $(r=1.1)$, where the redundancy factor r controls the amount of redundancy introduced during the transformation.


\subsection{Perplexity and Accuracy Results of 1–2 Bit Quantized Models}
The perplexity and zero-shot accuracy results of previous $1$-$2$ bit quantization methods and the proposed HBLLM are presented in Table\ref{tab:summary-table}. HBLLM consistently outperforms existing $1$-$2$ bit quantization techniques across all evaluation metrics. 

Specifically, HBLLM reduces the language modeling perplexity by $33\%$-$66\%$ compared to previous methods, while achieving substantial improvements in QA task accuracy, with relative gains ranging from $-0.73\%$ to $+11.3\%$. In our experiments, HBLLM slightly outperforms FrameQuant, a $2.2$-bit quantization method, and exhibits a particularly significant advantage on the LLaMA-3-8B model. Moreover, when compared with BiLLM and ARB-LLM\textsubscript{X}, HBLLM-col, demonstrates a clear advantage in both perplexity and accuracy, despite operating at comparable or lower bit-widths.
These results indicate that HBLLM effectively narrows the performance gap between quantized models and their Float16 counterparts, achieving $1.22\times$ to $2.48\times$ of the original perplexity and retaining $73.8$\%-$88.8$\% of the original QA accuracy.
\begin{table}[t]
	\centering
	\small
	\caption{Ablation study on LLaMA2-7B. Results are measured by perplexity, with final results highlighted in \textbf{bold}.}
	\label{tab:ablation-llama7b}
	
	\begin{subtable}{0.48\textwidth}
		\caption{Study of salient column selection criterion}
		\label{tab:ablation-a}
		\begin{tabular}{lccc}
			\toprule
			
			\cellcolor{gray!15} & \cellcolor{gray!15}\textbf{Selection} & \cellcolor{gray!15}& \cellcolor{gray!15} \\
			\cellcolor{gray!15}\multirow{-2}{*}{\textbf{Method}}& \cellcolor{gray!15}\textbf{criterion}&\cellcolor{gray!15}\multirow{-2}{*}{\textbf{Wiki2↓} }&\cellcolor{gray!15}\multirow{-2}{*}{\textbf{PTB↓}}\\
			\midrule
			\multirow{2}{*}{HBLLM-row} & $\ell_1$ & 10.78 & 143.7 \\
			& $\ell_2$ & \textbf{10.52} & \textbf{89.23} \\
			\hdashline
			
			\multirow{2}{*}{HBLLM-col} & \rule{0pt}{2.2ex}$\ell_1$ & 11.45 & 308.2 \\
			& $\ell_2$ & \textbf{11.33} & \textbf{150.6} \\
			\bottomrule
		\end{tabular}
	\end{subtable}
	\hfill
	\begin{subtable}{0.49\textwidth}
		\caption{Study of grouping granularity}
		\label{tab:ablation-b}
		\begin{tabular}{lccc}
			\toprule
			\cellcolor{gray!15} & \cellcolor{gray!15}\textbf{Group} & \cellcolor{gray!15}& \cellcolor{gray!15} \\
			\cellcolor{gray!15}\multirow{-2}{*}{\textbf{Method}}& \cellcolor{gray!15}\textbf{Partition}&\cellcolor{gray!15}\multirow{-2}{*}{\textbf{Wiki2↓} }&\cellcolor{gray!15}\multirow{-2}{*}{\textbf{PTB↓}}\\
			\midrule
			\multirow{2}{*}{HBLLM-row} & global & 16.32 & 1990 \\
			& row-wise & \textbf{11.08} & \textbf{95.58} \\
			\hdashline
			\multirow{2}{*}{HBLLM-col} & \rule{0pt}{2.2ex}global & 13.99 & 1546 \\
			& row-wise & \textbf{12.02} & \textbf{146.1} \\
			\bottomrule
		\end{tabular}
	\end{subtable}
	
	\vspace{0.08em} 
	\begin{subtable}{0.48\textwidth}
		\caption{Effectiveness of shared mean}
		\label{tab:ablation-c}
		\begin{tabular}{lccc}
			\toprule
			\cellcolor{gray!15} & \cellcolor{gray!15}\textbf{Shared} & \cellcolor{gray!15}& \cellcolor{gray!15} \\
			\cellcolor{gray!15}\multirow{-2}{*}{\textbf{Method}}& \cellcolor{gray!15}\textbf{mean}&\cellcolor{gray!15}\multirow{-2}{*}{\textbf{Wiki2↓} }&\cellcolor{gray!15}\multirow{-2}{*}{\textbf{PTB↓}}\\
			\midrule
			\multirow{2}{*}{HBLLM-row} & \ding{55} & 11.08 & 95.58 \\
			& \ding{51} & \textbf{10.52} & \textbf{89.23} \\
			\hdashline
			\multirow{2}{*}{HBLLM-col} & \rule{0pt}{2.2ex}\ding{55} & 12.02 & \textbf{146.1} \\
			& \ding{51} & \textbf{11.33} & 150.6 \\
			\bottomrule
		\end{tabular}
	\end{subtable}
	\hfill
	\begin{subtable}{0.5\textwidth}
		\caption{Study of partitioning candidates number}
		\label{tab:ablation-d}
		\begin{tabular}{lccc}
			\toprule
			\cellcolor{gray!15} & \cellcolor{gray!15}\textbf{Candidate} & \cellcolor{gray!15}& \cellcolor{gray!15} \\
			\cellcolor{gray!15}\multirow{-2}{*}{\textbf{Method}}& \cellcolor{gray!15}\textbf{number}&\cellcolor{gray!15}\multirow{-2}{*}{\textbf{Wiki2↓} }&\cellcolor{gray!15}\multirow{-2}{*}{\textbf{PTB↓}}\\ 
			\midrule
			\multirow{4}{*}{HBLLM-row} & 10 & 11.16 & 108.8 \\
			& 20 & 11.32 & 165.8 \\
			& 40 & \textbf{11.08} & \textbf{95.58} \\
			& 80 & 11.13 & 113.8 \\
			\bottomrule
		\end{tabular}
	\end{subtable}
	
\end{table}

\subsection{Ablation Study}

\textbf{Salient Column Selection Criterion.} To evaluate the impact of selection criteria in salient column screening on quantization effectiveness, we compare two strategies: the column $\ell_1$ norm and the column $\ell_2$ norm as significance indicators. Experimental results in Table \ref{tab:ablation-a} reveal that the column $\ell_2$ norm consistently achieves lower quantization error and superior performance in downstream tasks, indicating its greater effectiveness in capturing energy distribution across columns and enhancing quantization quality.

\textbf{Granularity of Group Quantization.} To explore the influence of grouping granularity on model performance, we compare global grouping with row-wise grouping strategies, evaluating both quantization error and perplexity, as shown in Table \ref{tab:ablation-b}. The results reveal that row-wise grouping significantly reduces quantization error and achieves lower perplexity compared to global grouping. This suggests that finer-grained row-wise partitioning better preserves local data fidelity, leading to improved quantized inference performance.

\textbf{Shared Mean Strategy.} Under the standard dual-partition quantization setting, we further explore a compression strategy that shares the quantization center across two partitions within each row. By unifying the mean for both partitions, the storage overhead of quantization coefficients can be significantly reduced. Experimental results in Table \ref{tab:ablation-c} demonstrate that the shared mean strategy even slightly reduces quantization error without degrading perplexity, verifying its effectiveness and practicality.

\textbf{Choice of Partitioning Number.} We investigate the impact of varying the number of partition candidates on final quantization performance under the row-wise grouping setting. Specifically, for each row, we generate partition candidates based on absolute value percentiles ranging from $10\%$ to $90\%$, and evaluate the corresponding quantization error and perplexity, as shown in Table \ref{tab:ablation-d}. Experimental results indicate that moderately increasing the number of partition candidates can effectively reduce quantization error and further lower perplexity, while excessive partitioning yields diminishing returns and increases computational cost. Consequently, we adopt 40 partition candidates as the default setting to balance performance and efficiency.

\begin{wraptable}{r}{0.5\textwidth}
		\caption{Time comparison between LLM binarization methods and our HBLLM on LLaMA-1 with different model sizes.}
		\label{tab:time_comparison}
		\begin{tabular}{lccc}
			\toprule
			\cellcolor{gray!15}\textbf{Method}       & \cellcolor{gray!15}\textbf{7B} & \cellcolor{gray!15}\textbf{13B} & \cellcolor{gray!15}\textbf{30B} \\
			\midrule
			BiLLM                 & 36min               & 71min                & 142min               \\
			ARB-LLM\textsubscript{x}   & 88min               & \ding{55}            &  \ding{55}            \\
			ARB-LLM\textsubscript{RC}  & 76min               & 119min               & 239min               \\
			PB-LLM                & 18min               & 29min                & 57min                \\
			FrameQuant            & 14min               & 22min                &  \ding{55}           \\
			HBLLM                 & 44min               & 98min                & 173min              \\
			\bottomrule
		\end{tabular}
\end{wraptable}

\subsection{Time and Memory Analysis}

\textbf{Time Comparison.} As a binary PTQ framework, HBLLM eliminates the need for finetuning. The introduction of Haar wavelet transforms requires additional computation during quantization, yet this overhead remains fully acceptable. As shown in Table \ref{tab:time_comparison}, HBLLM increases the quantization time by approximately $20\%$-$30\%$ compared to BiLLM across different model sizes. It is worth noting that ARB-LLM\textsubscript{X} and FrameQuant fail to complete quantization for LLaMA-$1$-13B and LLaMA-$1$-30B under the single-GPU-24 GB setting, while HBLLM successfully completes the process, demonstrating better scalability.

\begin{wraptable}{r}{0.5\textwidth}
	\caption{Memory comparison LLM binarization methods and our HBLLM on LLaMA-1 with different model sizes. }
	\label{tab:memory_comparison}
	\begin{tabularx}{0.5\textwidth}{lYY}
		\toprule
		\cellcolor{gray!15}\textbf{Method}       & \cellcolor{gray!15}\textbf{7B} & \cellcolor{gray!15}\textbf{13B} \\
		\midrule
		FP16                  & 13.48GB             & 26.03GB              \\
		BiLLM                 & 2.93GB              & 5.36GB               \\
		ARB-LLM\textsubscript{x}   & {3.23}GB              & {5.95}GB               \\
		ARB-LLM\textsubscript{RC}  & {2.83}GB              & {5.17}GB               \\
		PB-LLM                & 2.91GB              & 5.33GB               \\
		{FrameQuant}  & {11.36GB}  & {16.08GB}  \\
		HBLLM-row             & 3.09GB              & 5.89GB               \\
		HBLLM-col             & 2.67GB              & 5.06GB  \\
		\bottomrule            
	\end{tabularx}
\end{wraptable}

\textbf{Memory Comparison.} As shown in Table \ref{tab:memory_comparison}, HBLLM-col achieves better performance while occupying a storage size comparable to ARB-LLM. By employing a grouped shared-mean strategy, HBLLM improves compression efficiency without sacrificing performance. Specifically, HBLLM-col applies Haar transforms along the column dimension, such that only one grouped quantization operation is required per row on the transformed coefficients. Compared to HBLLM-row, this leads to reduced data fidelity but provides clear advantages in storage cost. The detailed storage calculation formulas can be found in the appendix {D}.

\subsection{Inference Latency Estimation}
{To evaluate the inference latency of HBLLM, we conduct an experiment that combines direct measurement with estimation. Due to there is no existing inference framework that fully supports the dequantization algorithm used in HBLLM, we test GEMV on layers from the OPT-175B model instead. The tests are run on an NVIDIA P100 GPU following the GPTQ benchmark setup [1].
Our estimation results show that the inference latency of HBLLM is approximately $31.8\%$ of the FP16 baseline inference time. For more details, please refer to the appendix G.}

\section{Conclusion}
We introduce a $1$-bit weight only quantization HBLLM, which applies Haar transform to BILLM pipeline. Besides quantifying the coefficients on frequence domain, HBLLM integrates two innovative structure-aware grouping strategies to enhance fidelity. 
Furthermore, HBLLM optimize storage efficiency. As a results, HBLLM outperforms SOTA QAT quantization methods of LLM at $1$-bit across different LLM families and tests. The current HBLLM supports only quantized dense models. Next, we will focus on the MoE PTQ algorithm.

\section*{Acknowledgements}
We thank all constructive comments from anonymous reviewers. This work is partially supported by the National Key Research and Development Program of China under Grant No.2023YFB3001704.
{
\small

\newpage
\section*{NeurIPS Paper Checklist}

The checklist is designed to encourage best practices for responsible machine learning research, addressing issues of reproducibility, transparency, research ethics, and societal impact. Do not remove the checklist: {\bf The papers not including the checklist will be desk rejected.} The checklist should follow the references and follow the (optional) supplemental material.  The checklist does NOT count towards the page
limit. 

Please read the checklist guidelines carefully for information on how to answer these questions. For each question in the checklist:
\begin{itemize}
    \item You should answer \answerYes{}, \answerNo{}, or \answerNA{}.
    \item \answerNA{} means either that the question is Not Applicable for that particular paper or the relevant information is Not Available.
    \item Please provide a short (1–2 sentence) justification right after your answer (even for NA). 
\end{itemize}

{\bf The checklist answers are an integral part of your paper submission.} They are visible to the reviewers, area chairs, senior area chairs, and ethics reviewers. You will be asked to also include it (after eventual revisions) with the final version of your paper, and its final version will be published with the paper.

The reviewers of your paper will be asked to use the checklist as one of the factors in their evaluation. While "\answerYes{}" is generally preferable to "\answerNo{}", it is perfectly acceptable to answer "\answerNo{}" provided a proper justification is given (e.g., "error bars are not reported because it would be too computationally expensive" or "we were unable to find the license for the dataset we used"). In general, answering "\answerNo{}" or "\answerNA{}" is not grounds for rejection. While the questions are phrased in a binary way, we acknowledge that the true answer is often more nuanced, so please just use your best judgment and write a justification to elaborate. All supporting evidence can appear either in the main paper or the supplemental material, provided in appendix. If you answer \answerYes{} to a question, in the justification please point to the section(s) where related material for the question can be found.

IMPORTANT, please:
\begin{itemize}
    \item {\bf Delete this instruction block, but keep the section heading ``NeurIPS Paper Checklist"},
    \item  {\bf Keep the checklist subsection headings, questions/answers and guidelines below.}
    \item {\bf Do not modify the questions and only use the provided macros for your answers}.
\end{itemize}


\begin{enumerate}

\item {\bf Claims}
    \item[] Question: Do the main claims made in the abstract and introduction accurately reflect the paper's contributions and scope?
    \item[] Answer:  \answerYes{} 
    \item[] Justification: The abstract and introduction clearly state our main contributions, which are supported by the theoretical and empirical results presented in the Section 3 and Section 4.

\item {\bf Limitations}
    \item[] Question: Does the paper discuss the limitations of the work performed by the authors?
    \item[] Answer: \answerYes{} 
    \item[] Justification: Please refer to Section 5.

\item {\bf Theory assumptions and proofs}
    \item[] Question: For each theoretical result, does the paper provide the full set of assumptions and a complete (and correct) proof?
    \item[] Answer: \answerYes{} 
    \item[] Justification: Please refer to Appendix.

    \item {\bf Experimental result reproducibility}
    \item[] Question: Does the paper fully disclose all the information needed to reproduce the main experimental results of the paper to the extent that it affects the main claims and/or conclusions of the paper (regardless of whether the code and data are provided or not)?
    \item[] Answer: \answerYes{} 
    \item[] Justification: Please refer to Section 4.

\item {\bf Open access to data and code}
    \item[] Question: Does the paper provide open access to the data and code, with sufficient instructions to faithfully reproduce the main experimental results, as described in supplemental material?
    \item[] Answer: \answerYes{}
    \item[] Justification: The code will be made publicly available upon acceptance of this paper. 

\item {\bf Experimental setting/details}
    \item[] Question: Does the paper specify all the training and test details (e.g., data splits, hyperparameters, how they were chosen, type of optimizer, etc.) necessary to understand the results?
    \item[] Answer: \answerYes{}
    \item[] Justification: Please refer to Section 4.1.

\item {\bf Experiment statistical significance}
    \item[] Question: Does the paper report error bars suitably and correctly defined or other appropriate information about the statistical significance of the experiments?
    \item[] Answer: \answerNo{} 
    \item[] Justification: Error bars are not reported because it would be too computationally expensive.

\item {\bf Experiments compute resources}
    \item[] Question: For each experiment, does the paper provide sufficient information on the computer resources (type of compute workers, memory, time of execution) needed to reproduce the experiments?
    \item[] Answer: \answerYes{}
    \item[] Justification: Quantization for models <30B was run on 4$\times$RTX 3090 (24GB), and for models $\geq$30B on A800-80GB. See Section 4.1.
    
\item {\bf Code of ethics}
    \item[] Question: Does the research conducted in the paper conform, in every respect, with the NeurIPS Code of Ethics \url{https://neurips.cc/public/EthicsGuidelines}?
    \item[] Answer: \answerYes{}
    \item[] Justification: We carefully read and follow the NeurIPS Code of Ethics.

\item {\bf Broader impacts}
    \item[] Question: Does the paper discuss both potential positive societal impacts and negative societal impacts of the work performed?
    \item[] Answer: \answerNo{}
    \item[] Justification:  The paper is purely fundamental research and does not involve social impact.
    
\item {\bf Safeguards}
    \item[] Question: Does the paper describe safeguards that have been put in place for responsible release of data or models that have a high risk for misuse (e.g., pretrained language models, image generators, or scraped datasets)?
    \item[] Answer: \answerNo{}
    \item[] Justification: This paper does not release new models or datasets.

\item {\bf Licenses for existing assets}
    \item[] Question: Are the creators or original owners of assets (e.g., code, data, models), used in the paper, properly credited and are the license and terms of use explicitly mentioned and properly respected?
    \item[] Answer: \answerYes{}
    \item[] Justification:  Please refer to Appendix.

\item {\bf New assets}
    \item[] Question: Are new assets introduced in the paper well documented and is the documentation provided alongside the assets?
    \item[] Answer: \answerNA{}
    \item[] Justification: The paper does not release new assets.

\item {\bf Crowdsourcing and research with human subjects}
    \item[] Question: For crowdsourcing experiments and research with human subjects, does the paper include the full text of instructions given to participants and screenshots, if applicable, as well as details about compensation (if any)? 
    \item[] Answer: \answerNA{}
    \item[] Justification: : The paper does not involve crowdsourcing nor research with human subjects.

\item {\bf Institutional review board (IRB) approvals or equivalent for research with human subjects}
    \item[] Question: Does the paper describe potential risks incurred by study participants, whether such risks were disclosed to the subjects, and whether Institutional Review Board (IRB) approvals (or an equivalent approval/review based on the requirements of your country or institution) were obtained?
    \item[] Answer: \answerNA{}
    \item[] Justification: The paper does not involve crowdsourcing nor research with human subjects.

\item {\bf Declaration of LLM usage}
    \item[] Question: Does the paper describe the usage of LLMs if it is an important, original, or non-standard component of the core methods in this research? Note that if the LLM is used only for writing, editing, or formatting purposes and does not impact the core methodology, scientific rigorousness, or originality of the research, declaration is not required.
    \item[] Answer: \answerNA{}
    \item[] Justification: LLMs were not involved as core components in the development of the method or results in this paper.

\end{enumerate}

\end{document}